\documentclass[nofootinbib,prd,aps,superscriptaddress,notitlepage]{revtex4-1}

\usepackage[utf8]{inputenc}
\usepackage{amsmath, bm}
\usepackage{amssymb,amsfonts}
\usepackage{hyperref}
\usepackage{graphicx}
\usepackage{grffile}

\begin{document}

\title{Understanding understanding: a renormalization group inspired model of (artificial) intelligence}
\author{A. Jakovac}
\affiliation{Wigner Research Centre for Physics, 1121 Budapest, Hungary}
\author{D. Berényi}
\affiliation{Wigner Research Centre for Physics, 1121 Budapest, Hungary}
\author{P. Posfay}
\affiliation{Wigner Research Centre for Physics, 1121 Budapest, Hungary}
\date{\today} 

\begin{abstract}
	This paper is about the meaning of understanding in scientific and in artificial intelligent systems. We give a mathematical definition of the understanding, where, contrary to the common wisdom, we define the probability space on the input set, and we treat the transformation made by an intelligent actor not as a loss of information, but instead a reorganization of the information in the framework of a new coordinate system. We introduce, following the ideas of physical renormalization group, the notions of relevant and irrelevant parameters, and discuss, how the different AI tasks can be interpreted along these concepts, and how the process of learning can be described. We show, how scientific understanding fits into this framework, and demonstrate, what is the difference between a scientific task and pattern recognition. We also introduce a measure of relevance, which is useful for performing lossy compression.
\end{abstract}

\maketitle

\section{Introduction}

To orient the reader we first draw up the structure of this paper. First, in this introductory section we try to motivate why is it important to have a comprehensive view about understanding, how it could help to make improvements in natural sciences and artificial intelligence (AI) systems. Next (Section \ref{sec:math}) we turn to the mathematical formulation of understanding, with the corresponding proofs and some examples from the field of AI. In the subsequent section (Section \ref{sec:science}) we examine, how understanding in natural sciences can be formulated to accommodate to these definitions. Here we also introduce the notion of the measure of relevance, which is important when we want to do a lossy compression of data. In Section \ref{sec:numrel} we try to identify the systems, where the traditional scientific method is the best, and where the AI methods can be expected to give a more reliable information about the functioning of the system. In Section \ref{sec:scratch} we examine, how in realistic time dependent systems the learning can be organized. Section \ref{sec:concl} is a summary of findings.

\subsection{Knowledge and understanding in natural sciences}

Knowledge and understanding is of central importance in human thinking, not a wonder that these concepts always were heavily debated in philosophy (epistemology \cite{epistemology}). In the natural sciences, on the other hand, these concepts traditionally had much simpler meaning. Simply stating, knowledge is all the facts that we can measure or prove in some way about the system under consideration, and understanding means that we have a model, best a mathematical model, that provides relations between the different facts. If we can purify our system such that it contains only the most basic facts, and all others we can compute from our equations: then we declare to understand our system fully.

This line of thought leads natural sciences to an analytic thinking. Indeed if we can reduce the system to its ingredients, and we can establish the way, how these elementary parts interact with each other, then we usually arrive at a simpler, more transparent system. Using this microscopic description, we can predict everything that happens in the complicated system. It is just a question of computing power.

In the endeavor to learn the smallest ingredients possible, we relentlessly hunted for the ultimate model, the Theory of Everything, the cradle of all understanding. So far we reached with considerable effort a spatial resolution of $\sim10^{-20}$m, and established the Standard Model \cite{SM}, that describes all of our measurements at this scale within measurable precision; even gravity, with a slight extension of the original model \cite{SHW}. We understand the Standard Model, we can do calculations in it, and, at least in principle we could calculate any process that is possible in the world. So from a practical point of view we already have the Theory of Everything at hand. All other is the question of computing power.

Computing power until the mid XX. century meant calculations on paper with a pencil. Since then, however, computers took over the most tedious part of the work, and did it with such an efficiency that exceeded the bravest dreams of our ancestors. But with the rise of the computers one may start to feel that we lost something from the clear understandability of our system. The two sources that play central role in this unsettling are the simulations of complicated (nonlinear) systems and the artificial intelligence networks.

\subsection{Challenges of understanding of complex system dynamics}

The treatment of complicated systems traditionally is done via perturbation theory if we use the paper and pencil method. There we know a simplified system that is solvable and in certain sense it is close to our complete system under consideration. Then we can try to use its results to approach the complicated system. But, as it turns out, this process has its limitations, and we can not reach beyond a certain limit, as perturbative series become divergent (e.g. QCD at small energies).

The only method that remains is the application of computers. But these provide us with pure bits and numbers, which do not tell us more, than as if they were stemming from a real world experiment.

So we stand and wonder, why we know all numbers (well, at least in principle), and still do not understand the reason. A prominent example is the existence of mass gap in Yang-Mills theories, which is a computer-established fact, but the reason is still one of the Millennium Prize Problems. But there is an ample number of examples, where we understand the underlying system much better than the emergent one: nuclear physics vs. QCD, molecular physics vs. quantum mechanics, protein folding vs. amino acids, or thinking vs. neurons.

There is another disturbing fact about understanding a scientific model. Traditionally the terms in a dynamic system are interpretable one-by-one, for example in a hydrodynamic model we can identify the term responsible for the pressure or shear viscosity of the fluid. But to approach a complicated model, we need many terms in the equation, so the dynamics is approximated by some, more-or-less motivated expansion. In this case the individual terms are just auxiliary quantities, they have no meaning one-by-one, only their cumulative effect matters. Completely different representations that yield the same dynamics, are equally good. This is, however, a disturbing phenomenon. We shall ask the question when a term in the dynamic evolution is interpretable, when is not; and how it can be conciliated with the understanding.

\subsection{Challenges of understanding in artificial intelligence}

The recent profound success of different Machine Learning (ML) approaches, in particular the deep neural networks (DNN) leaves a number of theoretical questions behind. Practically we may ask, why does DNN perform better than fixed ``scientific'' algorithms in certain fields like face recognition, or table games? Why a deep learning network is so effective? What are the questions that are worth to address with machine learning? How can we reduce the learning set? What is common in image classification and Newton equations? 

We see the technicalities. We know that a feed forward DNN is in fact a parametrizable function $y={\cal N}(x,W)$ which maps the $x$ input to $y$ output depending on $W$ parameters (weights) that modify the properties of the function. In a given architecture we can explicitly write up this function, and we can identify and read out the value of each parameter if necessary. But, although the program fulfills its job, it seems to intelligently ``understand'' the problems from, sometimes a huge number of examples, still\dots. Finally we just stand around and speculate what the heck it \emph{really} does.

There are several problems with neural networks. First, that they make errors unexpectedly. Also the best trained classificators are vulnerable to adversarial attacks, much more than a human. This is a signal that an AI "understands" the environment in a different way than we do.

Another challenge is that DNNs, when trained by different initial conditions, or shown examples in different orders, will exhibit different weights. The models, described by these DNNs, thus are ultimately different. Still, they describe the same reality with comparable accuracy. It seems that individual terms have no significance in a DNN, only what they describe together.

\subsection{A loose definition of understanding}

\begin{figure}[htbp]
  \centering
  \includegraphics[height=4cm]{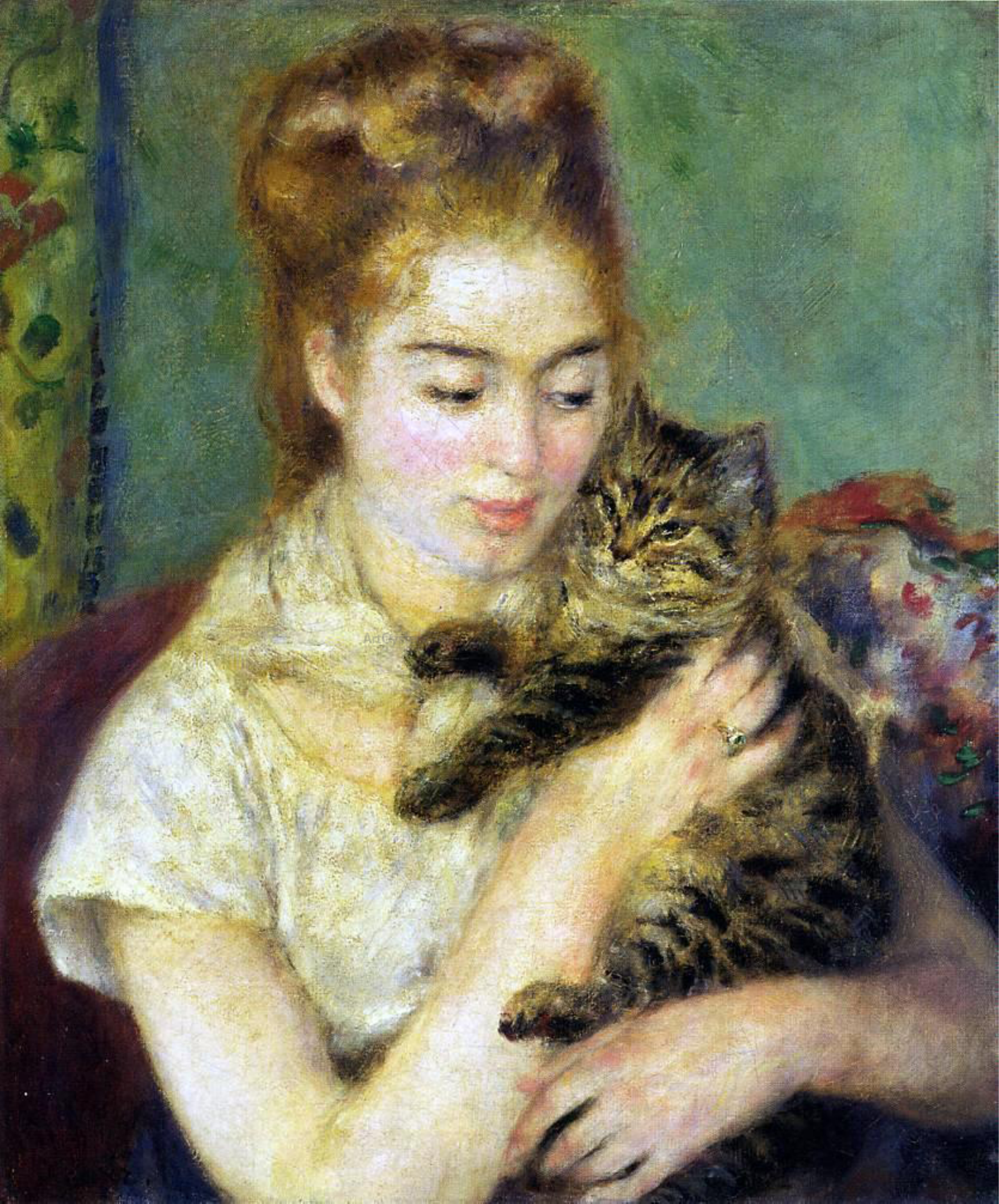}  
  \caption{Renoir: Woman with a cat}
  \label{fig:renoir}
\end{figure}

Motivated by these thoughts we approach understanding from a point of view that differs from the traditional natural science approach. Let us start with an example: try to describe a picture, say the one in Figure~\ref{fig:renoir}. In my computer it appears as a digitized image, with the resolution of 1Mpixel. Putting these pixels after each other the content of the original picture can be recognized (to the given resolution). But we know that the original picture is painted, so originally there are no pixels there but paintbrush strokes instead. If we apply these strokes one after another, the image will be recognizable, again within the given resolution provided by the brush width (an art forger does it almost perfectly). But we can also try and describe the picture starting with the objects we recognize in the image.
We may start to tell "this is an image where there is a woman with a tabby cat, the woman wears a yellow blouse and a scarf, the blouse is crimped at the armpits, the woman has a bunned blond hair \dots''. If we wish, we can go into any small details, finally potentially describing the content of the image in the precision of a computer pixel. To do so, we need to mention one million details, much more than how many useful notions we have, but in principle it is feasible. 

This example demonstrates, that if we have only a single object to describe, there is no preferred narrative, all descriptions are equally good, each of them reflects a different point of view how we contemplate the image. For a computer expert the important is the storage, and transformations of the image. They need it in a digitized form, they cannot do anything with the representation in human notions (cat, woman, etc.). On the other hand someone who just want to enjoy the aesthetic picture, need not care about the elementary manifestation of it (if it is digitized or painted).

But if we have a collection of images, then we already have a singled out view. This view is that grabs,  in a sense, the common features of all the images. For example we can have Renoir paintings, or cat images, or images where the top left pixel is brown. For our above example all are true, but a Renoir painting usually does not portray a cat, and in most cat pictures the top left pixel is not brown, and so on. For a collection of images we have a best description.

If we know what is common in the images, a lot of questions can be answered easily. For example if we show a random image, we can decide whether it is a part of our collection or not. If we collected all the cat images, than a random cat image is part of our collection, a random dog image is not. If we are asked to show an example of our collection, it is enough to draw a cat image, since it is certainly an element of all cat images. We may realize that these are exactly the tasks an AI network is asked for.

The lesson of these thoughts is twofold. First we see that what we can understand is a collection of images; in a mathematical sense we can understand a subset of some large set. Second, we can verify that we understand a collection, if we can accomplish all usual AI tasks (classification, regression, lossless data compression, encoding, etc.) easily. These thoughts will be recasted in a mathematical formulation in the present work.

\subsection{Studies in the literature}

There are very nice existing approaches to the question of understanding in computer science. Predecessor is the learning theory \cite{learningtheory, LTWiki}, where mathematical definition is used to build up in the framework of Probably Approximately Correct model of the system. From our point of view the closest approach is representation learning \cite{RL}, where the goal is to find a good representation of the input data in order to facilitate the optimization of a ML architecture. Concepts are introduced recently in \cite{SCAN}, and in Ref. \cite{disentang} the authors deal with disentangling representations with the help of symmetry groups.

In physics the main ideas come from the (exact) renormalization group \cite{ERG}. These ideas were also used in computer science, c.f. Boltzmann Machines \cite{BM}, or applications of renormalization group \cite{RGinAI}.

\section{Mathematical approach to understanding}
\label{sec:math}

In this section we aim to write down the mathematical definition of understanding, give the necessary proofs, and demonstrate that we indeed defined the concept of understanding. For that we first need some preparatory steps.

\subsection{The basic definitions}

Let us have a set $X$, this will be the embedding set of our subset that we want to understand. We will assume that $|X|<\infty$ is a finite set. Example is $X=\{$set of 1Mpixel color images$\}$, where each pixel can have $256^3 = 2^{24} = 16.7$ million colors. Altogether $|X|=(2^{24})^{10^6}\approx 10^{7\,224\,720}$. A vastly large, but still a finite set.

We will approach this set in two different ways. First we will \emph{coordinatize} it in some information conserving way. This means that we look for bijective mappings $\xi: X \mapsto B^N$, where $B\subset\mathbb R$ is some finite set of real numbers and $N = \log|X|/\log|B|$ (assumed to be an integer); we mostly will use $B=\{0,1\}$  binary set, which corresponds to a bit of a pixel in an image. The cardinality of all coordinatizations is $|X|! \approx \sqrt{2\pi |X|} (|X|/e)^{|X|}$ from the Stirling formula (the same as the number of permutations). These coordinatizations are the mappings provided by neural networks, extended to a bijective map. Technically we cannot achieve all the bijections in a DNN, only the ones that can be parametrized by a $|W|$ sized parameter space, where $W$ is the set of all weights. If a parameter can have $n$ different states, then there are $n^{|W|}$ possibilities, this is the number of different DNNs. Taking into account that the number of parameters is of order of a million, the parameters itself may be a 4- or 8-byte long number, we have $2^{64000000}$ different permutations, which is a tiny subset of all possible bijections ($n^{|W|}$ compared to $|X|!$).

The other way we approach the set of inputs is a probabilistic one. As we have discussed, for a single input there is no preferred coordinate system. If we have a subset of inputs, then it already singles out (not uniquely) the best representation. For this we should speak about the distribution of the coordinates, this will be defined below.

We will establish a probability space $(X, F_X, \mathbb P_X)$, where the sample space is the \emph{input} space, the event space $F_X = 2^X$ is its power set (i.e. the set of all subsets), and the probability measure $\mathbb P_X : F_X\to [0,1]$ is the uniform discrete probability measure $\mathbb P_X(C) = |C|/|X|$. In this probability space the $\xi_i:X\to B$ coordinates can be considered as random variables. For example $X$ can be the set of 1MPixel images, then $F_X$ is the set of all subset of 1Mpixel images, like the cat images or the landscape images.

We remark that in the usual approach for classification, we define the probability space over the $Y$ \emph{output} space of the neural network, and the $\xi_i(x) \in [0,1]$ normalized coordinates play the role of the probability measure for any given $x\in X$: there we consider $(Y, F_Y, \xi(x))$.

The coordinates, being random variables, allow to speak about their (conditional) distributions. Let us consider a $C\subset X$ subset and an index set $I=\{i_1,\dots,i_a\} \subseteq \{1,2,\dots N\}$. We define $p_I^{(C)}(\xi=\sigma)$ where $\sigma\in B^N$, the (joint) distribution of the coordinates $I$ over $C$ as the conditional probability distribution
\begin{equation}
  p_I^{(C)}(\xi=\sigma) = \mathbb P_X(\{x\in X| \xi_i(x) = \sigma_i,\, \forall i\in I\} \;|\;C) =   \frac1{|C|} \sum\limits_{y\in C} \delta(\xi_{i_1}(y)=\sigma_{i_1}) \dots \delta(\xi_{i_a}(y)=\sigma_{i_a}),
\end{equation}
where $\sigma = (\sigma_i)_{i\in I} \in B^{|I|}$ and $\delta(i=j)=1$ if $i=j$ and $0$ if $i\neq j$ (Kronecker delta). Putting into words, we count the cases from a fixed subset of the input, when the value of each chosen coordinate is equal to the specified values. For $I=\{1,\dots,N\}$ we will omit the lower index: $p^{(C)}(\xi=\sigma)$, for $I=\{i\}$ we denote $p_{\{i\}}^{(C)}(\xi=\sigma) =p_i^{(C)}(\xi=\sigma)$.

Note that the conditional expected value of a $Q:X\to \mathbb R$ (measurable) function (a.k.a. a random variable) over this subset reads
\begin{equation}
  \mathbb E(Q \,|\,C ) =\frac1{|C|} \sum_{y\in C}Q(y) = \sum_{\sigma \in B^N} Q\circ \xi^{-1}(\sigma)\,p^{(C)}(\xi=\sigma).
\end{equation}
This is a coordinatization independent concept. In particular the expected value of the product of coordinates $\mathbb E(\xi_{i_1}\dots \xi_{i_a} \,|\,C)$ is called ($a$-point) correlation function (note that $B\subset \mathbb{R}$).

The $i$th coordinate is said to be \emph{independent} from the others, if 
\begin{equation}
  p^{(C)}(\xi=\sigma) = p_i^{(C)}(\xi = \sigma) p_{\{1,\dots,N\}-\{i\}}^{(C)}(\xi=\sigma).
\end{equation}
In words, the $ith$ distribution factorizes. In this case any correlation function reads
\begin{equation}
  \mathbb E(\xi_i \xi_{i_1}\dots \xi_{i_a}\,|\; C) = \mathbb E(\xi_i\,|\, C)\cdot \mathbb E(\xi_{i_1}\dots \xi_{i_a}\, |\, C).
\end{equation}
In the binary case knowing all correlation functions is equivalent to knowing the distribution. In particular if all correlation functions factorize, then the coordinates are all independent: this is an effective check for the independence of coordinates in the binary case.

The value of a coordinate is \emph{deterministic} over $C\subset X$, if for the complete studied $C$ subset it takes a fixed value. A deterministic coordinate is always independent from the others, because
\begin{equation}
	p^{(C)}(\xi=\sigma) = \delta(\sigma_i=v)\, p_{\{1,\dots,N\}-\{i\}}^{(C)}(\xi=\sigma),
\end{equation}
where $v$ is the fixed value of the first coordinate.

The \emph{uniform distribution} is where the random variable takes all possible values with equal probability. If the $i$th coordinate is uniform, then we have
\begin{equation}
	p_i^{(C)}(\xi=\sigma) = \frac1{|B|}.
\end{equation}

Let us see some examples with $B=\{0,1\}$.
\begin{itemize}
\item If $C = \{c\}$ has only a single element, then its coordinates are all deterministic (and as such, independent), and their distribution over $C$ is
\begin{equation}
	p^{(\{c\})}(\xi=\sigma) = \prod_i \delta(\sigma_i=c_i).
\end{equation}
To generate this image we have to set the value of all the coordinates exactly.
\item Take $C=X$, i.e. all images. Then the distribution of the coordinates is
\begin{equation}
	p^{(X)}(\xi=\sigma) = \frac1{|X|},
\end{equation}
uniform, and so for the $i$th pixel
\begin{equation}
  p_i^{(X)}(\xi=\sigma) = \frac12.
\end{equation}
The expected value of each coordinate is
\begin{equation}
 \mathbb E(\xi_i\,|\,X) = \frac12.
\end{equation}
Since the product $\xi_{i_1} \xi_{i_2}\dots \xi_{i_a} = 1$ only, if all pixels have value 1, therefore
\begin{equation}
  \mathbb E(\xi_{i_1} \xi_{i_2}\dots \xi_{i_a}\,|\,X) = \frac1{2^a} = 
  \mathbb E(\xi_{i_1}\,|\, X)\dots \mathbb E(\xi_{i_a}\,|\, X).
\end{equation}
This means that the coordinates are uncorrelated, independently of the choice of the coordinatization.

Since the coordinates are uncorrelated, if we choose independently a value for all pixels, then we generate a random image in $X$.
\item Although in these examples the coordinates of the input were uncorrelated, this nice property goes away when we have more complicated examples. Let us begin with a very simple example, and consider three bits. Then
\begin{equation}
  X=\{ 000,001,010,011,100,101,110,111\},\qquad |X|=8.
\end{equation}
Now choose a subset of images
\begin{equation}
  C= \{ 001,010,100,111\},\qquad |C|=4.
\end{equation}
Considering the first bit it takes the value $0,0,1,1,$ on subsequent elements of $C$. This means for its distribution
\begin{equation}
  p_1^{(C)}(\xi=\sigma) = \frac12.
\end{equation}
This is true for all other coordinates, too. Consequently $\mathbb E(\xi_i\,|\,C)=1/2$ for all $i=1,2,3$. 

For the 2-point correlation functions $\xi_1\xi_2 =1$ only if both are 1, this occurs only once ($111$). Therefore
\begin{equation}
  \mathbb E (\xi_1 \xi_2\,|\,C) = \frac14 =\mathbb E(\xi_1\,|\,C)\mathbb E(\xi_2\,|\,C),
\end{equation}
and similarly for all other pairs, meaning that the 2-point correlation functions factorize.

But these distributions are not independent, because the 3-point correlation function does not factorize
\begin{equation}
  \mathbb E (\xi_1 \xi_2\xi_3\,|\,C) = \frac14 \neq \mathbb E(\xi_1\,|\,C)\mathbb E(\xi_2\,|\,C)\mathbb E(\xi_3\,|\,C) = \frac18.
\end{equation}
Indeed, if we forget about the correlation of the coordinates, and try to pick up random values for $\xi_1,\xi_2$ and $\xi_3$, we generate images from $X$, and not from $C$.
\end{itemize}

\subsection{Mathematical definition of understanding}

The lesson from the previous analysis is that in a subset of images the pixels lose their independence, and hidden, often high-order correlations appear. This makes it impossible to reconstruct $C$ from the knowledge of the pixelwise distribution functions.

How nice would it be to find a representation that ``undoes'' this correlation! We could choose the representative coordinates freely in order to produce an example of the subset! As we will see, in fact this is what we heuristically called ``understanding''. 

\begin{list}{{\bf def.:}}{}
\item A $\xi:X\to B^N$ bijection (coordinatization) is called a \emph{complete model} of a subset $C\subset X$, if the coordinates are independent, and the $p_i^{(C)}(\xi=\sigma)$ distributions are either deterministic, or have uniform distribution for all $i\in\{1,2,\dots, N\}$.
\end{list}

We can define similar notions for a set of disjoint sets:
\begin{list}{{\bf def}}{}
\item  A $\xi:X\to B^N$ bijection (coordinatization) is called a \emph{common complete model} of a $\{C_a, C\}$ system, where $C_a\subset X$ subsets are pairwise disjoint and $C=\cup_a C^a \subset X$, if the coordinates are independent, and they are either deterministic, or have uniform distribution for all $C_a$ and $C$.
\end{list}

Having physical correspondance in mind we will define the relavant and irrelavant coordinates of the common complete model:

\begin{list}{{\bf def.:}}{}
\item \emph{overall relevant coordinates} are the deterministic coordinates from the point of view of $C$.
\item \emph{partially relevant coordinates} are the those independent coordinates that are either deterministic or uniformly distributed for all $C_a$ and $C$, and there exists at least one subset, where it is deterministic.
\item \emph{irrelevant coordinates} are those independent coordinates that have uniform distribution for all $C_i$.
\end{list}

It is clear that the complete model of $C$ is not unique: by combining the relevant coordinates among themselves, or combining the irrelevant coordinates among themselves (maintaining uniform distribution) results in another complete model.

And now the definition of understanding:
\begin{list}{{\bf def.:}}{}
\item We \emph{understand} a subset $C\subset X$ if we can provide a complete model for $C$
\end{list}
and
\begin{list}{{\bf def.:}}{}
\item We \emph{understand} a $\{C_a, C\}$ system, where $C_a\subset X$ subsets are pairwise disjoint and $C=\cup_a C^a \subset X$, if we can provide a common complete model.
\end{list}

Below we first prove that there always exists a complete model for all $C$ subsets, and a common complete model for all sets of pairwise disjoint sets. Next, we will show examples that support why the above definition really can be thought as understanding.

\subsection{Proof of the existence of a common complete model}

We give here an explicit construction of the common complete model.

Let us have a collection of pairwise disjoint subsets of the basic set $C_a\subset X$, and $C_a\cap C_b =0,\;\forall a\neq b$. We will construct a $\xi: X\to \{0,1\}^N$ coordinatization where all coordinates are independent, and either deterministic, or uniformly distributed between $0$ and $1$.

The easiest is the construction when $|C_a| = 2^{N_a}$, $|C|=|\cup_a C_a| = 2^{N_C}$ and $|X|=2^N$ with $N_a, N_C, N\in\bm N$. We may assume without loss of generality that $N_1\ge N_2\ge\dots \ge N_a$. Then we make an indexed list:
\begin{equation}
	(\mathrm{elements\ of\ }C_1, \mathrm{elements\ of\ }C_2,\dots \mathrm{elements\ of\ }C_a, \mathrm{elements\ of\ }X-C).
\end{equation}

\begin{list}{{\bf Statement:}}{}
\item The index of the above list forms a complete common model.
\end{list}
\begin{list}{{\emph{Proof:}}}{}
\item The mapping that maps the index to the element is bijection, since all elements occur, and occur only once. The starting index of $C_i$ subset is
\[ 2^{N_1}+2^{N_2}+\dots+2^{N_{i-1}} = K\,2^{N_i},\qquad K = 2^{N_1-N_i}+2^{N_2-N_i}+\dots+2^{N_{i-1}-N_i}.
\]
$K$ is an integer, because $N_j\ge N_i$ for $j<i$. The $C_i$ subset ends at $(K+1)\,2^{N_i}-1$, because it has $2^{N_i}$ elements. Therefore in the binary representation of the indices of the elements of $C_i$ the first $N-N_i$ bits are constant. Taking into account only the last $N_i$ bits, then all possibilities occur, and so these bits are independent and uniformly distributed.
\end{list}

In other words, the first $N-N_i$ coordinates are relevant, the last $N_i$ coordinates are irrelevant from the point of view of $C_i$. From the point of view of $C$ the first $N-N_C$ coordinates are relevant, the rest is irrelevant.

\textbf{As an example} consider $X$ as a 4-bit system
\begin{equation}
 X = \{0000, 0001, 0010, 0011,\dots 1100, 1101, 1110, 1111\},\quad |X|=16,
\end{equation}
and
\begin{equation}
 C_1 = \{0010, 0101, 1000, 1100\},\qquad C_2 = \{0110, 1111\}\qquad C_3 = \{0111, 1001\}.
\end{equation}
To have a common complete model we list the elements of the subsets in decreasing order in the cardinality of the subsets. This means:
\begin{equation}
\begin{array}{rcl}
	\mathrm{index} & = & \{0000, 0001, 0010, 0011, 0100, 0101, 0110, 0111, \dots\} \cr	
	\mathrm{elements} & = & \{0010, 0101, 1000, 1100, 0110, 1111, 0111, 1001, \dots\}. \cr
\end{array}
\end{equation}
As we see, the first 4 items in the "elements" list come from $C_1$, then come the subsets $C_2$ and $C_3$, respectively. The corresponding coordinatization maps $0010\mapsto0000$, $0101\mapsto0001$, \dots. Now we can identify the relevant and irrelevant coordinates:
\begin{itemize}
\item The elements of $C_1$ are the ones for that the first two coordinate bits are $00$. From the point of view of $C_1$ the first two bits are relevant, the second two are irrelevant. The irrelevant bits are independent, and uniformly distributed.
\item The elements of $C_2$ are the ones for that the first three coordinate bits are $010$. From the point of view of $C_2$ the first three bits are relevant, the last is irrelevant. The irrelevant bit is uniformly distributed within $C_2$.
\item The elements of $C_3$ are the ones for that the first three coordinate bits are $011$. From the point of view of $C_3$ the first three bits are relevant, the last is irrelevant. The irrelevant bit is uniformly distributed within $C_3$.
\item From the point of view of $C = C_1\cup C_2\cup C_3$ the relevant bit is the first one (it is deterministically 0 within $C$), the three remaining bits are irrelevant: independent and uniformly distributed.
\end{itemize}
As it should be evident, the above construction does not depend on the actual elements of the chosen subsets.

The construction becomes somewhat more complicated if the number of elements are not powers of 2. Clearly if $|X|$ is not a power of 2, then there cannot be a bijection between $X$ and $\{0,1\}^N$ for any $N$. To overcome this difficulty, we extend all $C_a$ with new dummy elements, $C$ with new dummy subsets, and $X$ again with dummy elements to achieve a power of 2 cardinality for each. Then the previous construction works through, we have a common complete model for the extended sets. The inverse of this bijection therefore goes from the coordinates to the extended set, not all coordinate combination yields element in $X$. Practically these elements should be neglected from the output (a.k.a. accept-reject method). Thus, extended with dummy elements we can ensure the existence of a common complete model.

In practice we do not necessarily extend the subsets, and as a consequence there can be some coordinates that are neither relevant, nor irrelevant. This spoils the beautiful mathematics, but in practice these few bits do not really count, when we have, say, one million bits (coordinates) together. 

\subsection{Continuous approximation}

Instead of having binary mappings, usually one works with floating point representation. The relation of the two is not too complicated, since we can treat the binary coordinates as binary figures of a floating point number, with a given precision. Having $n_b$ coordinates $\sigma_i\,i=1\dots n_b$, the corresponding floating number will be
\begin{equation}
	x = \sum\limits_{i=1}^{n_b} \frac{\sigma_i}{2^i},\qquad x\in [0,1[.
\end{equation}
This provides a bijection between floating point numbers with $n_b$ significant binary figures and the $n_b$ binary coordinates.

This construction can be performed for the input image as well as for the coordinates of the common complete model. In the latter case we should collect either relevant or irrelevant binary coordinates. If all the binary coordinates are relevant, then the floating point representative will be relevant, too, with deterministic value within the given subsets. If the binary coordinates are all irrelevant, having the same probability for taking 0 or 1 value, then the corresponding floating representative will be irrelevant, too, i.e. independent from all the other coordinates, moreover, it will have a uniform distribution in the $[0,1)$ range.

Knowing that this is possible, we may directly start with a real number representation of the input $X\sim \mathbb R^N$, and look for the common complete model for a collection of (continuous) subsets $C_a\subset X$ where all coordinates will be independent, and either relevant (i.e. having deterministic values) or irrelevant (i.e. having uniform distribution) for all the subsets as well as for their union.
This coordinate system will then be a curvilinear system from the point of view of the representation of the input.

This representation is useful, if we can expect some continuity in the singled out subsets; or in more general terms if we can expect some differential manifold structure for the input as well as for the singled out subsets. This is true for a lot of cases: for example an image in pixel representation probably belongs to the same classification class as its neighbor where we changed some pixels, thus the subsets are continuous.

A continuous curvilinear coordinatization of a differential manifold is a comfortable approach, but it may have some unwanted side effects. For example it may be singular in some points of the input, when it ceases to be a bijection, or it can not be extended to the complete manifold with a single map. Although some of these problems may have significance in the original problem, we must keep in mind that the continuous representation is just an approximation, and the true images and coordinates are all discrete.

\subsection{Artificial Intelligence tasks performed through a complete model}

Having a complete common model, we can easily solve all the AI tasks. We demonstrate in this subsection, how.

\subsubsection{Classification}

Classification means that we have to tell apart the elements of the disjoint sets $C_1,\dots C_a$, knowing that the items are in the union set $x\in C = \cup_a C_a$. This last condition tells us that the relevant bits of $C$ are given; the irrelevant bits do not influence in which subset is the element (that is why it is called irrelevant). So we have to inspect the \emph{partially relevant} coordinates, they are uniquely determine the $C_i$ subset.

In the above example we want to find the class of $0110$ image. Its coordinates $\xi(0110) = 0100$. The first bit is always zero for all elements of $C$. We have to examine the next ones. We know that the relevant leading bits (apart from the leading zero) of $C_1$ is $0$, of $C_2$ is $10$, for $C_3$ is $11$. The given item has $10$, thus it belongs to $C_2$.

We note that the first bit tells us whether the classification task is feasible at all by inspecting the bits that are relevant for the union set. If these do not agree with the deterministic values of the union set, then we can tell that the input is not an element of any class. An outlier is detected and identified as such! 

\subsubsection{Regression}

The task of regression is to tell the parameters of a function when the function values are noisy. For example we have a set of $\{(x,y)\}$ point pairs that are around an exponential function $y=ae^{bx}$, and then we have to tell the corresponding $a$ and $b$ values.

Regression can be treated as a classification problem. Here the $C_{a,b}$ sets are noisy exponentials around a smooth function $ae^{bx}$. To ensure disjointness, we associate a set of point pairs $\{x_i, y_i\}$ to that $C_{a,b}$, where the probability to have this pattern is the largest, according to some noise model.

Since $a$ and $b$ are computer-stored numbers, we have a finite number of disjoint $C_{a,b}$ subsets. Having a complete common model, we can decide that a given point pair set $\{x_i, y_i\}$ belongs to which $C_{a,b}$, by simply inspecting the partially relevant coordinates. We also can tell, if the noisy function does not belong to any of the classes. 

\subsubsection{Decoding}

Decoding is a task that we ask from the AI to produce a random element in $C$. We know that the relevant bits $\sigma_{relevant} =\xi_{relevant}(x)$ are constant for $x\in C$. Therefore to produce en element in $C$ we apply
\begin{equation}
  \xi^{-1}(\sigma_{relevant}, \sigma_{irrelevant}=\mathrm{random\, uniform}) \in C.
\end{equation}
Moreover, since the distribution of irrelevant coordinates is uniform, the above construction provides all elements of $C$ with equal probability.

\subsubsection{Data compression}

From the decoding task we see that to fully characterize an element of $C$ it is enough to remember the irrelevant coordinates, since we already know the (constant) value of the relevant ones. Therefore compression is a map
\begin{equation}
  U:x\mapsto\xi_{irrelevant}(x).
\end{equation}
This is a lossless compression, since we can undo it with
\begin{equation}
  x= \xi^{-1}(\sigma_{relevant}, U(x)),\qquad \forall x\in C.
\end{equation}

\section{Model building in natural sciences, and measure of relevance}
\label{sec:science}

The above mathematical model can be actualized to the understanding phenomena in natural sciences. The basic set $X$ is the states or processes of the complete environment. To characterize it, we perform all kind of measurement, we collect all possible facts about the world. There are, of course, infinitely many facts, and not all of them are easily accessible, but in principle "all the facts" forms the basis of the original characterization of the world. Usually, in a scientific approach not the value of the measurement, but the measurement operation (measurement operator) is given. However, in  any actual environment this is translated to real numbers.

From $X$ we single out $C^{(M)}_a\subset X$ subsets that make up a "phenomenon", and which we want to understand. These subsets themselves come from a measurement $M$ (which can be a collection of several elementary measurements), and the $a$ index runs through the set of possible outcomes of that measurement. That is, we collect into $C^{(M)}_a$ all the states/processes of the world, where the $M$ measurement yields the result $a$.

\begin{figure}[htbp]
\begin{center}
   \includegraphics[height=2.5cm]{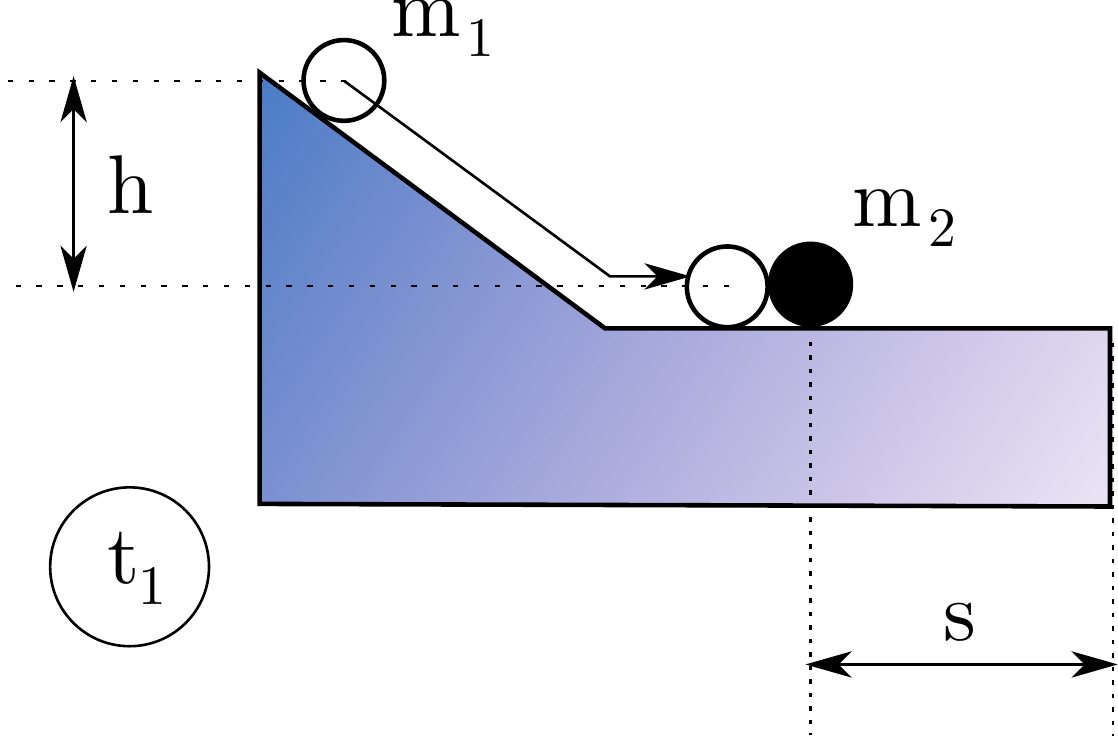}\qquad 
   \includegraphics[height=2.2cm]{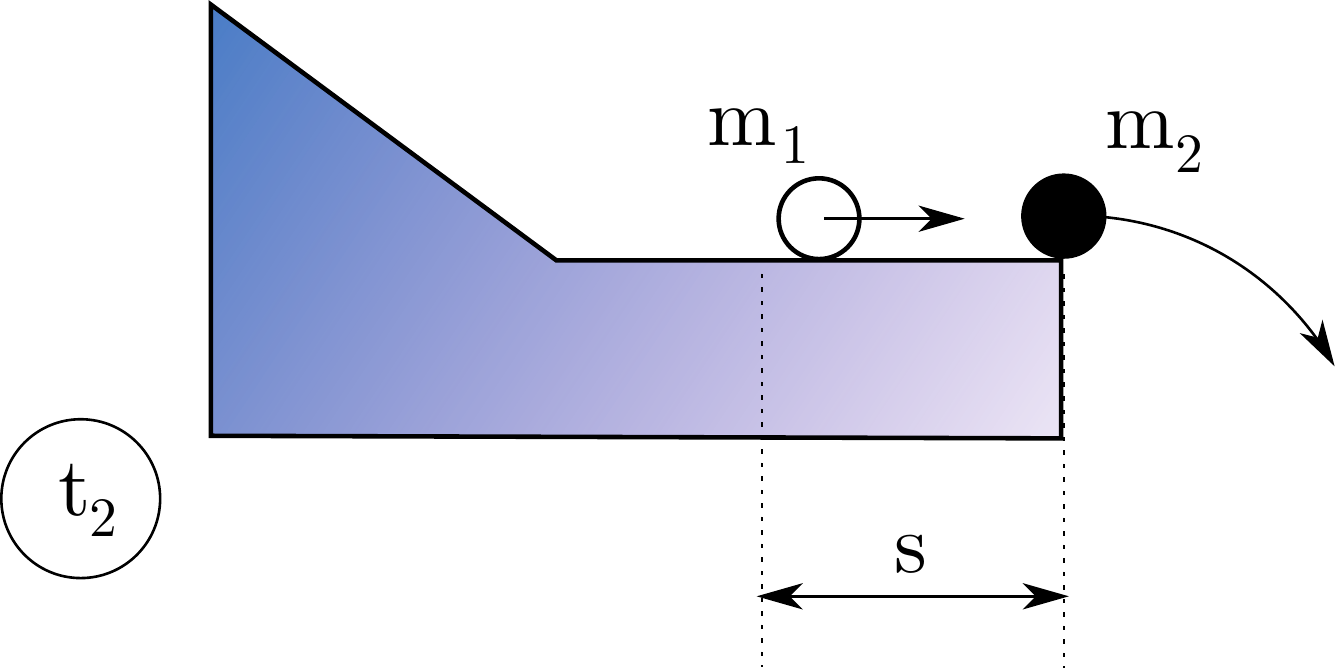}
\end{center}
   \caption{The physical phenomena we want to understand: a ball rolling down a slope, hits another ball which subsequently fall off the table.}
   \label{fig:slope}
\end{figure}

To give an actual example, let us consider a ball rolling down in a slope from some height, then colliding with another ball, which finally falls off the table (c.f. Figure~\ref{fig:slope}).
The phenomenon we want to understand is the time difference $\Delta t$ between the collision and the fall off: this measurement is symbolized by $M$. We chose $a=\Delta t$ for labeling the subsets, where we collect all experiments that result in the same measured value.

According to our earlier definitions, to understand the phenomena we need to have a common complete model for all $C^{(M)}_a$ subsets. So assume we have found it, and try to interpret the coordinates. There are the overall relevant coordinates of all the experiments, i.e. relevant coordinates of $C^{(M)}=\cup_a C^{(M)}_a$. These represent the common framework of all experiments, for example that we have a slope, a straight trough to lead the ball, or that we use solid balls. There are the overall irrelevant quantities: these are the ones that do not influence the results of the experiments, because anyhow we chose their value, we remain in the same subset $C^{(M)}_a$: these are for example the color of the balls, or the person of the experimenter. Thus the most important quantities are the partially relevant ones, that are irrelevant for $C^{(M)}$, but relevant for $C^{(M)}_a$. These quantities have a fixed value in each experiment, but changing their value will change $a$, i.e. the result of the experiment.

To be concrete, in the simple experiment described above we know, what are the quantities that influence the result: 
\begin{equation}
\Delta t = t_2-t_1 = \frac s { 2 \sqrt{2gh}  } ( 1+ \frac{m_2}{m_1} ).
\end{equation}
That means that the time difference depends only on some measured facts, like the height of the slope, the distance of the colliding spot to the edge of the table, etc.. All other details do not matter.

A scientific model deals with a lot of experiments of the same kind, and tries to find the relevant coordinates that influence the values of the measurements. For example we can deal with the mechanics of rigid bodies, we study their motion, rotation in different environments. The rigidity of the bodies is an immutable part of all experiments, in our language this provides the overall relevant coordinates. Put another way these present the defining framework of the discipline of mechanics of rigid bodies. Within this framework we shall find those phenomena, or coordinates that are relevant from the point of view of single measurements. The mechanics of rigid bodies uses coordinates like center of mass position, velocity, axis of rotation, angular velocity, angular acceleration, etc. There are some 10-20 coordinates that are important.

Once we know the complete set of relevant coordinates, i.e. the "physical" quantities, we can be sure that any measurement can be expressed through them as a general function $f(x,t,v,\omega,a,\beta, \Theta, \dots)$. By carefully analyzing a lot of experiments we discover that certain coordinates always appear together, and based on this observation we can associate a unit of measurement to all of them. This reduces the number of the possible combinations essentially.

An important step if we realize that the relevant coordinates we use to characterize the system are not necessary independent, i.e. we have an overcomplete basis. Then we can reveal relations between the coordinates to make them complete. These relations are the (natural) laws, for example $F=ma$ Newton's law.

The recognition of the "law" that the shape of the body and the moment of inertia are not independent, and there are formulae that can express $\Theta$ from the shape, may initiate an even further step: we may introduce coordinates that are not directly observable in the system, but these are the true independent coordinates of the measurements. In this way in mechanics we arrive at point mechanics, which helps to extend all the findings of mechanics of rigid bodies to mechanics of elastic bodies and even to fluid mechanics.

\subsection{Noise, measure of relevance and renormalization}

All what was said above needs some refinement. In reality the output of measurements usually form a near continuous ensemble, so we can interpret all measurements as $X\to \mathbb R$ mappings. But then we in vain fix the "physical" parameters of a measurement, we never measure the same values. In mechanics it is interpreted that the measurements are corrupted by noise, in statistical physics we speak about fluctuations. But noise or fluctuation are part of the system, so what we measure is a well defined process (at most we do not measure exclusively that we really want). Since we labeled the subsets by the measured values of our singled out $M$ experiment, if we measure enough, and with ultimate precision, all subsets will consist of just a single element. Indeed, all changes in the process influence the result of the measurement in some, maybe extremely tiny, extent.

For a solution we must define our measurements more cautiously, and associate a finite resolution to them. This can be done in several ways; one possibility is that we consider the result only up to a finite precision $p$, keeping just a given number of significant figures $n\sim p$. Then the index $a$ in $C_a$ will refer to intervals (characterized by an integer, for example), and thus $|C_a|>1$, and the above analysis goes through.

In this way we can define a family of measurements that differ in the precision $p$ we treat the result. As $p$ decreases, $|C_a|$s increase, and the number of irrelevant quantities grow. At zero precision we collect all measurements into one set, so $C=X$, and all coordinates are irrelevant. Each coordinate has a $p_i$ value associated to it, where it becomes irrelevant: this may serve as a measure of relevance.

Another way is to define the singled out measurement(s) $M$ that first average out the state or process before we measure, on a \emph{scale} $k$ (i.e. the spatial or temporal interval where we do the smearing is $1/k$), and keep the result up to a finite (fixed) precision. Then again we obtain a $C_a^{(k)}$ series of subsets, where $|C_a^{(k)}|$ will grow by lowering the scale.

In physics the changing scale is used to be taken into account in a more quantitative way. Here we usually use the energy measurement (Hamiltonian) to represent a phenomenon, so $M(k)=H(k)$. If we have a basis of measurements $\xi_i$, then $H$ can be expanded in this base
\begin{equation}
 H(k) = \sum_a g_a(k) \xi_a,
\end{equation}
where the $g_a$ quantities are called couplings. For example the kinetic energy is $H = p^2/2m$, here the basis element is the momentum squared measurement, the coupling is the inverse mass. The above formula says that if we change the energy measurement definition, then, in a given basis, we change the couplings. This phenomenon is called "renormalization" or "running coupling", respectively.

The Hamiltonian has a double role: on the one hand it defines the energy measurement, on the other hand it governs the dynamical processes. In particular it also describes the dynamics of the fluctuations or noise. Thus when we lower the scale (increase the regime of averaging), then the fluctuations at the given scale will give a contribution to the change of the couplings in a definite way, derived from the Hamiltonian itself. This leads to the self-consistent equation:
\begin{equation}
 \label{eq:RGbeta}
 \frac{d g_a}{dk} = \beta_a(k,g).
\end{equation}
The $\beta_a$ beta-functions are characteristic to the studied system.

The above equations \eqref{eq:RGbeta} are known as renormalization group equations, and they are central concepts in statistical physics, with extensive literature. Here we only treat a simplified case. We linearize the right hand side, using that $\beta(k, 0)=0$ (which follows from the fact that if there were no energy at a given scale $k$, then there will not be at another $k$, too). We find
\begin{equation}
 \frac{d g_a}{dk} = \sum_b \beta_{ab}(k) g_b + \mathcal{O}(g^2),
\end{equation}
where $\beta_{ab} = \partial \beta_a/\partial g_b|_{g=0}$. In physically sensible systems this matrix can be diagonalized, and the eigenvalues are real numbers:
\begin{equation}
	\beta_{ab}(k) = \sum_n \lambda_n(k) u_{an}(k) v^*_{bn}(k),\qquad
	\delta_{ab} = \sum_n u_{an}(k) v^*_{bn}(k),
\end{equation}
where $\lambda_n\in \mathbb R$ are the eigenvalues, $u_n$ and $v_n$ are the left and right eigenvectors, respectively. 

The simplification we consider here will be that the eigenvectors of $\beta_{ab}(k)$ are independent of the scale. Then for $h_n = v^*_{an}g_a$ we find
\begin{equation}
	\frac{d h_n}{dk} = \lambda_n(k) h_n,
\end{equation}
meaning that these $h_n$ couplings evolve independently. There are two types of behavior: for $\lambda_n(k)<0$ the solution grows with decreasing $k$, for $\lambda_n(k)>0$ it decreases with decreasing scale.

We can redefine the Hamiltonian with these couplings as
\begin{equation}
	H(k) = \sum_n h_n(k) \eta_n, \qquad\mathrm{where}\quad \eta_n =\sum_b v^*_{bn}\xi_b.
\end{equation}
By decreasing the scale, i.e. making the fluctuations less and less profound, we see that certain terms remain important, exactly those, where $\lambda_n(k)<0$: these terms are therefore relevant from the point of view of energy measurement. The ones with $\lambda_n(k)>0$ will decrease, so these become irrelevant. As measure of relevance we can use the actual value of the couplings $h_n$.

We should add that in real systems $\beta_a(k,g)$ can be very complicated. For example it can happen that $\lambda_n$ does not remain consistently positive or negative. Then the given coordinate changes its relevance. For example the temperature of a system is not a relevant quantity for nuclear physics under normal conditions, but with lowering the scale, i.e. considering larger and larger bodies, sooner or later it becomes relevant, and forms the basis of thermodynamics. In the meantime the physical properties of nuclei, which is important for the nuclear physics, become irrelevant from the point of view of a macroscopic body.

\subsection{Symmetries, and the uniqueness of the coordinate representations}

Another issue worth to discuss is how unique is the relevant-irrelevant basis, and, in particular, how well defined are the terms in a Hamiltonian (or in general in a dynamic system).

The first point we have to pin down is that in physical systems there are quantities that have the same physical significance. For example our space is three dimensional, so if we want to coordinatize a position, we have to give three numbers. But the actual values are not fixed, because they vary with the choice of the coordinate system. As a consequence, the two functions
\begin{equation}
 f_1(x) = x_1,\qquad f_2(x) = a_1 x_1 + a_2 x_2 + a_3 x_3
\end{equation}
may describe the same reality, only in a different coordinate system.

In fact the transformation between the coordinates can be used to help to construct models where we expect independence of some coordinate transformations. For example, if we do not have a preferred direction in the space, then the energy of this system must depend only on the scalar product of spatial vectors.

How can be this observation incorporated into the choice of relevant-irrelevant basis?  If we have a $C\subset X$ subset in a binary system, it singles out the relevant and irrelevant bits, but their order is arbitrary. If we form from the bits floating numbers, then the permutation of relevant/irrelevant bits changes the value of these numbers, while keeping their relevance or irrelevance intact.

In the continuum, if we have independent deterministic, or uniform distributed random variables, then we can freely redefine them, provided we maintain this property. Therefore there is a big freedom in defining the relevant-irrelevant basis.

The previous subsection restricted even more what we expect from a good coordinatization: a nearby measurement, for example that differs only in setting the scale, should yield a similar basis. According to this restriction we can combine only those coordinates that have the same (or at least similar) measure of relevance.

If we have a few number of relevant quantities, then these restrictions single out almost uniquely the relevant part of the basis that is worth to use. In particular the Hamiltonian of a system with a few terms is practically uniquely defined. Then all of its terms can be identified, and measured one-by-one, in most cases we also name them (like mass, charge, velocity, momentum, etc.). When, however, the number of the relevant terms grow to large, we lose this property, just because all the parameters have a similar role, and they can be reshuffled among themselves. Practically a system with of order 20-30 relevant terms with similar relevancy is not uniquely defined, we can redefine the coordinates, still maintaining the same performance.

So the question whether the coordinates have an individual meaning, a "personality", depends on how many coordinates are there with a similar relevancy. That how many relevant coordinates do we have, is studied in the next section.

\section{How many relevant coordinates do we have?}
\label{sec:numrel}

As we have seen in the simple physical example above, a physical system has a few relevant quantities. So few, that we even name them (like masses, distance, height, velocity, etc.), and according to their role we can associate a unit of measure to them (like kg, m, m/s etc.).

On the other hand relevant quantities are the clue of image recognition as well. But how many relevant parameters have a set of images? To answer this question, we can use our earlier result, that for lossless compression we need to store only the values of the irrelevant coordinates, since the relevant ones are the same within a given subset.

In order to compare the scientific models and images we take the same input set, and we choose a subset either from a physical measurement, or from a collection of images. The input set is a binary image where each pixel can have a value $0$ or $1$.

\begin{figure}[htbp]
\begin{center}
	\includegraphics[height=3cm]{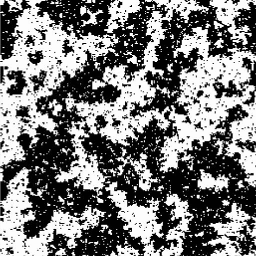}\qquad
	\includegraphics[height=3cm]{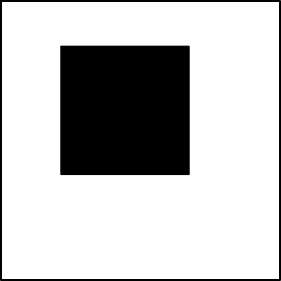}
	
	a.) \hspace*{3cm} b.)
	\vspace*{-1em}
\end{center}
\caption{Binary image from the subset generated by a.) Ising fluctuations, and b.)  black square on white background}
\end{figure}

The physical system will be the Ising model, where the Hamiltonian comes from adding correlation functions of pixels with different weights. In the physical measurements we average large-scale correlation functions (called infrared physics) with weights $e^{-\beta H}$ where $H$ is the Hamiltonian, and $\beta$ is a parameter. In this system there are three relevant quantities, the $\beta$ parameter, the one where we count the number of pairs with non-equal values (i.e. $0-1$ or $1-0$), and the one where we add the number of 1-s. All the other quantities are irrelevant.

For the subset of the images we interpret $0$ as white and $1$ as black color, and we consider a single black square on a white background. To compress this image we need to remember the coordinates of top-left corner, and the length of the edge of the square. Knowing these parameters, we can reconstruct the original image exactly. In contrary, knowing the 4 relevant quantities of our ball experiment yields a rather poor description of the real experiment, we do not know, who made it, where, what was the color and the material of the ball, etc. 

This means that we have altogether 3  \emph{irrelevant} coordinates in the black-and-white square images! If we have a 256x256 binary image, then it can be described by $65536$ bits. From it the 3 coordinates use $24$ bits, that is we have $65512$ relevant and $24$ irrelevant bits.

In more complicated examples the number of irrelevant bits also grow. For example if we take a 100x100 color image depicting at most 10 rectangles, it can be described by 486 irrelevant bits, which is still small as compared to the number of relevant bits ($\approx 240000$). For even more complicated cases we can assume that the number of both the relevant and the irrelevant bits are large.

The remarkable in this line of thought is that the relation of the numbers of the relevant/irrelevant parameters are opposite in image recognition and in physics problems. In physics problems we have a few relevant parameters, in image recognition we have a large number of relevant parameters. 

As we have learned in the last subsection, if we have a few relevant quantities, then they are more or less uniquely defineable, and they have their individual character. The same is true, if we have a few irrelevant quantities: then the irrelevants can be identified uniquely (c.f. our example with the black square). But neither is the case in a general image recognition task: the number of the relevant as well as the irrelevant quantities are large, many of them with similar relevancy. Therefore they do not matter one-by-one, only their cumulative effect is important. This explains the experience that different DNNs with vastly different weights may have a very similar performance.

It is also plausible that between the two extremes, the physics of simple systems and the plain black square image, the realistic systems form a more-or-less continuous ensemble. Mechanics, electrodynamics, spin systems, etc. are examples with the fewest relevant parameters. More complicated are the hydrodynamic and elastic models, but very soon we arrive to models where the number of relevant parameters is very hard to tell. For example we do not know all the relevant parameters of nuclear physics, despite of some 60 years of thorough study. In meteorology, chemistry or biology the situation gets more and more complicated.

All this means that if we have the ambition to describe a more complicated system precisely, we need to apply the methods of image recognition in order to determine the value of the relevant parameters. Their Hamiltonian, or equations of motion must be determined by neural networks, or similar intelligent ways. This also means that we have to rethink the scientific strategy that we used so far to approach these systems, as well as the belief in a fully interpretable Hamiltonian.

\section{Learning from scratch and the eureka experience}
\label{sec:scratch}

When we start learning, we do not yet know the complete subset that we need to understand. Let us examine what happens when we start to build up a set element by element. For simplicity we will speak about image recognition, but what is to be said applies to the scientific understanding, too.

We should start with the fact that for a smaller set there are more relevant parameters than for a larger set. Starting with a single image, all tiny information seems to be relevant, and there is no preferred coordinate system. When we perceive more and more images, the relevant and irrelevant quantities become more and more evident. Having $2^n$ images at hand in a class, we have $n$ irrelevant bits. 

To the contrary of the neural network learning methods, in reality the order of perception of new images is important, as well as the typical timescale of their variation. Since the observation of new images implies the appearance of new irrelevant coordinates, we can label the irrelevant coordinates with the typical time during which they change their value: this label corresponds to the "versatility" of this coordinate, and it is a good proxy for the measure of relevance.  

If we consider an actor, who perceives images in their whole lifetime, there are different types of coordinates. There are the ones that never change: these are the relevant parameters of the existing world itself. Clearly, an actor does not need to remember these, just because they never change. There are other coordinates that change very fast. These again are not worth to remember, because these do not influence the reality too much. The only coordinates that matter are the ones with a appreciable amplitude and a reasonable change rate.

This means that usually we do not use a complete model, we have to keep only those coordinates that are less irrelevant. Putting another way, we apply a lossy compression, which uses as an organizing principle the measure of relevance. What loss is tolareble, is a fine tuning question. If we omit too much, then we may miss crucial details (like in the grass lurking panther). Keeping to much details may hinder the brain identifying the really important effects (like in autism).

In supervised learning we explicitly tell the system which coordinates are the most relevant by training the outputs for each available input. This method has the advantage that we can use our well established coordinate system for a computation task, but also has disadvantages, because we can train only a small number of relevant coordinates, due to the lack of annotated examples.

Thus the solution should be to use a large number of unannotated inputs, and train the coordinates according to the measure of relevance, extracted for example from the time variability of that coordinate. Then we can not transfer our knowledge to the computer, but as a compensation it will have its own system of notions.

In more complicated systems the variability in time is not necessarily a good clue to assess the relevance of a parameter. Then we have to change more and more to a self-consistent definition, that a parameter is relevant, if it influences the value of other relevant parameters in the future. Since this is a self-consistent definition, one may sometimes stuck in a false parametrization.

It can happen that some coordinates seem to be relevant, but in reality they are not. This failure can come from bad data sampling, where by chance we overrepresent some features. Then we think that the given parameter influences the future significantly, and we are sometimes reluctant to change our model. Superstition or (false) conspiracy theories are examples of this case.

Another possibility is when we omit a parameter as irrelevant, although it has important effect to the future. Then we experience that we can just poorly describe the future values of the relevant parameters, and in this case one is tempted to say that there is an unexplorable randomness in the system (a frequent explanation for mispredicitons in the financial markets). If we believe in a logical explanation, and we are keen to find the true reason, we may find that some apparently irrelevant effects are responsible for the mispredictions. This is the "eureka" effect, or the catharsis of understanding.

But it is certainly true that this higher level learning is very difficult, and it is very hard to asses, how well a given coordinatization performs. The multitude of worldviews in humanity reflect the complexity of this problem.

\section{Conclusions}
\label{sec:concl}

In this work we tried to give a novel view on understanding, especially in natural sciences, including artificial intelligence. The main idea is that we can understand a collection of processes or states, which form a subset of all possible processes/states. Heuristically understanding means to reveal the common features of the subsets that identify them uniquely. Technically we have shown that a subset singles out a coordinatization, albeit not uniquely, where each coordinates are independent over the subset, and are either relevant (i.e. have deterministic value) or irrelevant (i.e. have uniform distribution over the complete domain).

This construction of relevant/irrelevant coordinates can be extended to several pairwise disjoint subsets, too. As a result we can easily solve all AI related tasks like classification, regression, lossless compression or decoding, by a simple inspection of the relevant or irrelevant coordinates. So, technically, we can answer all questions easily.

This formalism fits to the artificial intelligence approach, and it can be actualized to be applicable to scientific understanding. There we select a phenomenon characterized by a measurement $M$ (which can be also a bunch of elementary measurements), and sort the events according to their results of this measurement. These form the necessary subsets, and then we can follow the general procedure for understanding.

Considering a single measurement (yielding a set of subsets), the relevant and irrelevant coordinates are equally important. But in reality there are more and less important coordinates, mathematically corresponding to a measure of relevance. To define this we need to consider several "nearby" measurements, for example by changing the precision, or by averaging over a spatial or temporal range, characterized by a $k$ parameter. From the so-defined $M(k)$ measurements we can read out the relevance of a coordinate in several ways (like associating the $k$ value where a given coordinate becomes irrelevant, or considering the renormalization group running, or observing the typical time variation scale of the coordinate). With a measure of relevance a lossy compression becomes available.

The relevant-irrelevant coordinatization is not unique, although the independence of the coordinates, the uniform distribution of the irrelevant ones, and the associated measure of relevance restricts it pretty much. When we have a few relevant or irrelevant coordinates, this results in definite coordinates, which we can identify, study and name. In physical models usually this is the case, and we can then understand the relevant effects term by term. The same is true for geometric images, where geometry puts so strict constraint on the image that only a few irrelevant coordinates remain. In general, however, bot the number of relevant and irrelevant coordinates are so abundant, that they can not be defined uniquely. Then the individual coordinates have no meaning, and only their cumulative effect is descriptive. This also have an impact on the parametrization of the corresponding model: while with few relevant quantities we can determine the value of the individual parameters independently, in a complicated model we can only train them in bundle, and there are several equivalent parametrizations performing the same task (c.f. trained neural networks).

\section*{Acknowledgment}

The authors acknowledge useful discussions with T.S. Biro, D. Nagy, G Orbán. This work is supported by the Hungarian Research Fund NKFIH (OTKA) under
contracts No. K123815.


\begin{thebibliography}{0}
\bibitem{epistemology} \url{https://plato.stanford.edu/entries/epistemology/}
\bibitem{SM} \url{https://en.wikipedia.org/wiki/Standard_Model}
\bibitem{SHW} M. Shaposhnikov and Ch. Wetterich, "Asymptotic safety of gravity and the Higgs boson mass", Phys.Lett.B 683 (2010) 196-200, [0912.0208 [hep-th]]
\bibitem{learningtheory} {Osherson D.N., Stob M., Weinstein S.}, "Systems That Learn: An Introduction to Learning Theory for Cognitive and Computer Scientists", MIT, 1990.
\bibitem{LTWiki} \url{https://en.wikipedia.org/wiki/Computational_learning_theory}
\bibitem{RL} Y. Bengio, A. Courville and Pascal Vincent, "Representation Learning: A Review and New Perspectives", eprint: arXiv: 1206.5538 (cs.LG)
\bibitem{SCAN} I. Higgins, N. Sonnerat, L. Matthey, A. Pal, Ch. P. Burgess, M. Bosnjak, M. Shanahan, M. Botvinick, D. Hassabis, A. Lerchner, "SCAN: Learning Hierarchical Compositional Visual Concepts", https://arxiv.org/abs/1707.03389
\bibitem{disentang} I. Higgins, D. Amos, D. Pfau, S. Racaniere, L. Matthey, D. Rezende, A. Lerchner, "Towards a Definition of Disentangled Representations",  	https://arxiv.org/abs/1812.02230 [cs.LG] 
\bibitem{ERG} \url{https://en.wikipedia.org/wiki/Renormalization_group}
\bibitem{BM} \url{https://en.wikipedia.org/wiki/Boltzmann_machine}
\bibitem{RGinAI} P. Mehta and D. J. Schwab. “An exact mapping between the Variational Renormalization Group and Deep Learning”, arXiv: 1410.3831; H. Lin and M. Tegmark, "Why does deep and cheap learning work so well?", http://arxiv.org/abs/1608.08225
\end{thebibliography}
\end{document}